\documentclass[lettersize,journal]{IEEEtran}
\usepackage{amsmath,amsfonts}
\usepackage{algorithmic}
\usepackage{algorithm}
\usepackage{array}
\usepackage[caption=false,font=normalsize,labelfont=sf,textfont=sf]{subfig}
\usepackage{textcomp}
\usepackage[normalem]{ulem}
\useunder{\uline}{\ul}{}
\usepackage{stfloats}
\usepackage{multirow}
\usepackage{url}
\usepackage{verbatim}
\usepackage{graphicx}
\usepackage{cite}
\begin{document}

\title{A New Class Biorthogonal Spline Wavelet for Image Edge Detection}

\author{Dujuan Zhou\IEEEauthorrefmark{1}, Zizhao Yuan\IEEEauthorrefmark{1}
\thanks{\IEEEauthorrefmark{1}These authors contributed equally to this work.}
}

% The paper headers
\markboth{Journal of \LaTeX\ Class Files,~Vol.~14, No.~8, August~2021}%
{Shell \MakeLowercase{\textit{et al.}}: A Sample Article Using IEEEtran.cls for IEEE Journals}

\maketitle

\begin{abstract}
Spline wavelets have shown favorable characteristics for localizing in both time and frequency. In this paper, we propose a new biorthogonal cubic special spline wavelet (BCSSW), based on the Cohen--Daubechies--Feauveau wavelet construction method and the cubic special spline algorithm. BCSSW has better properties in compact support, symmetry, and frequency domain characteristics. However, current mainstream detection operators usually ignore the uncertain representation of regional pixels and global structures. To solve these problems, we propose a structural uncertainty-aware and multi-structure operator fusion detection algorithm (EDBSW) based on a new BCSSW spline wavelet. By constructing a spline wavelet that efficiently handles edge effects, we utilize structural uncertainty-aware modulus maxima to detect highly uncertain edge samples. The proposed wavelet detection operator utilizes the multi-structure morphological operator and fusion reconstruction strategy to effectively address anti-noise processing and edge information of different frequencies. Numerous experiments have demonstrated its excellent performance in reducing noise and capturing edge structure details.
\end{abstract}


\section{Introduction}
\IEEEPARstart{E}{dge} is a key feature of an image that is important for many vision tasks, including object detection, image segmentation, object recognition, tracking, and 3D reconstruction. The purpose of edge detection is to extract clear and continuous contour information from the target scene, providing simplified scene and object-of-interest information for advanced visual tasks. The detection effect is mainly affected by image noise, brightness, and contrast, which can result in false or broken edges. Therefore, it is crucial to design an effective detection operator that can mitigate these issues. Researchers investigate various methods for detecting operators, primarily for denoising, edge sharpening, continuity, and detail preservation. Edge detection methods are typically classified as traditional operator detection, wavelet transform-based detection, and neural operator (NO) detection. In the following section, we first review the current research on detection operators. \par
Previously proposed conventional operators primarily rely on variations in luminance and gradient to identify edges. The Canny \cite{ref-journal2} operator combines the gradient and non-maximum suppression methods to obtain refined edges. The Prewitt \cite{ref-journal3}, Sobel \cite{ref-thesis1}, Roberts \cite{ref-thesis2}, and Laplacian operators \cite{ref-journal18} use convolution kernels to filter the image and identify edges in the direction of the gradient. However, these operators are sensitive to noise and have low detection accuracy, making it difficult to perform adaptive edge detection. Tian et al. \cite{ref-journal4} combined the weighted kernel paradigm minimization and Sobel operator to improve the noise robustness capability of the Sobel operator and achieve better detection results. Mittal et al. \cite{ref-journal5} proposed the Canny optimization algorithm, which simulates triple thresholding and efficiently carries out edge detection while improving edge continuity. Isar et al. \cite{ref-journal6} proposed optimizing the Canny operator by introducing a two-stage denoising system in the wavelet transform domain to improve its robustness to noise. \par
Extracting edge features that are fully adapted to complex backgrounds remains a significant challenge due to limitations in image feature representation and differences in time-frequency domain analysis. Wavelet transform has gained significant attention from researchers due to its powerful multi-scale decomposition capability and spatial domain processing. The wavelet transform is commonly used in various visual tasks, such as image enhancement \cite{ref-proceeding2, ref-journal7}, denoising \cite{ref-journal8, ref-journal15}. For edge detection, wavelet-based detection operators \cite{ref-journal19, ref-journal20, ref-journal21, ref-proceeding6} typically combine wavelet transform and modulus maxima. Inspired by Wavelet modulus operators, Gu et al. \cite{ref-journal9} proposed an improved wavelet mode-maximization algorithm that enhances edge contours by fusing light intensity and polarized light. You et al. \cite{ref-journal10} further optimized the algorithm by incorporating OTSU thresholding to detect edges in complex background images. Morphology-based edge detection is effective in reducing noise, particularly edge noise, due to its rational structural element design. Shui et al. \cite{ref-journal11} proposed an edge detector that is resistant to impulse noise, based on anisotropic morphological directional derivatives, to achieve competitive edge results in both noiseless and Gaussian noise scenarios. Yin et al. \cite{ref-proceeding3} proposed a multi-scale, multi-directional approach using structural elements and a Mahalanobis distance-weighted detection operator to extract smoother, more detailed, and noise-resistant edges.
In recent years, there has been significant progress in machine learning (ML) methods. Dollar et al. \cite{ref-journal12} proposed a structured learning method for random decision forests with the aim of building fast and accurate edge detectors. Hallman et al. \cite{ref-proceeding4} proposed directed random forests based on simplified feature representations for linear boundary detection and edge sharpening of image blocks. Neural operators are often based on data-driven approaches, where they learn edge feature representations of labels to recognize edges more accurately. However, their performance heavily relies on well-labeled data \cite{ref-proceeding5} and they still suffer from coarser edges, poor noise robustness, and difficulty in retaining detailed information about object structure. More importantly, few researchers have studied the impact of different wavelets on the performance and effectiveness of edge detection algorithms. Therefore, in this work, we mainly propose a new, effective spline wavelet detection operator. This detection operator mainly includes structural uncertainty-aware modulus maxima and multi-structural anti-noise morphology. Experimental results show that the proposed spline wavelet detection operator has better anti-noise performance than other wavelets and operators, and can detect more complete and continuous edge information. Overall, our contributions are as follows: \par

\begin{itemize}
\item We propose a new and effective BCSSW for edge detection based on the CDF wavelet construction method and the cubic special spline algorithm, which provides better image smoothing and noise suppression, resulting in a high-quality low and high-frequency prior for detection compared to other wavelets.
\item We propose a novel BCSSW edge detector (EDBSW) based on the fusion of structural uncertainty perception modulus maxima and anti-noise operators. Inspired by uncertainty-aware detectors, we introduce low-frequency structural uncertainty perception into a wavelet transform-based detection operator for the first time. 
\item To our knowledge, we are the first to compare the effectiveness of different wavelets in edge detection with the proposed BCSSW quantitatively and qualitatively. We evaluate the proposed detection operator by substituting the filter parameters of various wavelets in the operator. The experimental results demonstrate that the proposed spline wavelet detection algorithm can excel in MSE and PSNR metrics compared to Haar and other wavelets.

\end{itemize}

\section{Methodology}
In this section, we presented the proposed BCSSW spline wavelet and edge detection algorithms, which contain (a) derivation of new spline wavelet low-high-pass filter coefficients, (b) modulus maxima detection algorithms based on structural uncertainty perception, (c) design and implementation of the multi-structural anti-noise operator, (d) fusion strategies for morphological refinement and reconstruction. \par

\begin{figure*}[ht]
\centering
\includegraphics[scale=0.35]{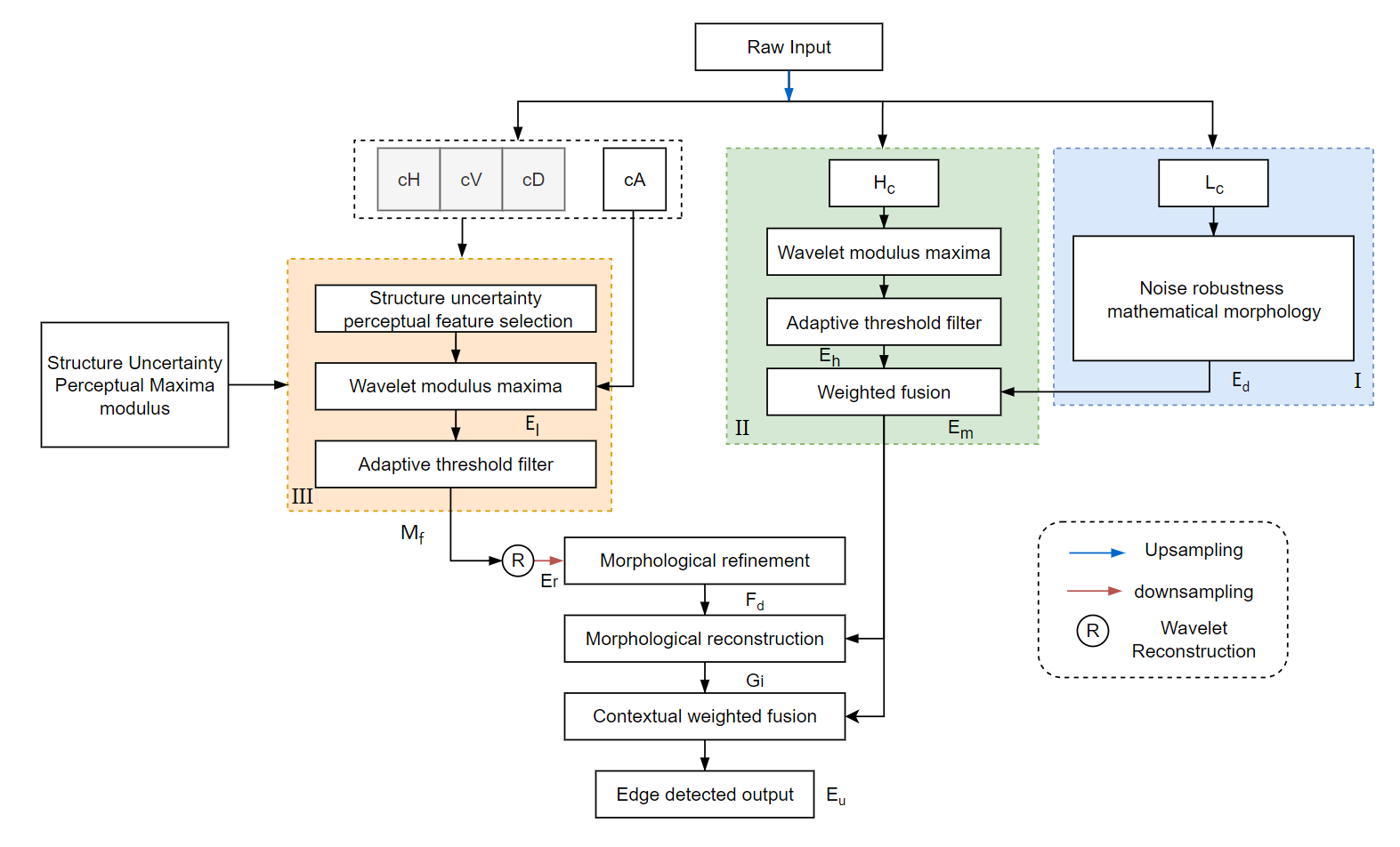}
\caption{The diagram of the proposed EDBSW.}
\label{fig1}
\end{figure*}

Overall, the diagram of the proposed new spline wavelet edge detection algorithm is shown in Fig.~\ref{fig1}. For the input image $I$ is upsampled and decomposed using the spline wavelet to obtain its low-pass and high-pass components. The first branch detects the low-frequency component $cA$ by applying a multi-structure anti-noise morphological operator to obtain the low-frequency feature map $E_{d}$. The high-pass component $H_{c}$ and low-pass component $L_{c}$ are then filtered sequentially using wavelet-based modulus maxima and adaptive threshold to obtain $E_{h}$ and $E_{l}$ in the second branch. These components are then weighted and fused to obtain the mask $E_{m}$. The third branch applies the high-frequency components $cH$, $cV$, and $cD$ to the structural uncertainty-aware modulus maxima to obtain structurally aware edge feature representations $CH^{'}$, $CV^{'}$, and $CD^{'}$. After the adaptive threshold filter, the high-frequency edge feature component $M_{f}$ undergoes wavelet reconstruction and downsampling to produce $E_{r}$. $E_{r}$ is then subjected to morphological refinement to obtain $G_{i}$. Eventually, $G_{i}$ and $E_{m}$ are fused by morphological reconstruction and context-weighted fusion of the original $E_{m}$ to obtain the final detailed edge image, $E_{u}$. The following section describes the specific algorithmic implementations and related parameter derivations of each branch in detail.

\subsection{A New Biorthogonal Cubic Special Spline Wavelet (BCSSW)}

\subsubsection{Cubic Special Spline Algorithm}
In our work, BCSSW is based on the cubic special spline algorithm proposed by \cite{ref-journal13}. They proposed a novel spline algorithm and provided various representations of cubic splines with different compact supports. In this paper, we selected one of the cubic splines with the smallest compact support, as shown in (\ref{eq:NCBW01}), and on the basis of this spline, we derived a new class of spline wavelet algorithms following the CDF method of wavelet construction.

\begin{equation}
\label{eq:NCBW01}
\begin{aligned}
S(t) = &\frac{451}{3}\beta_3(t)-\frac{256}{3}(\beta_3(t-\frac{1}{16})+\beta_3(t+\frac{1}{16}))+ \\
& \frac{64}{3}(\beta_3(t-\frac{1}{8})+\beta_3(t+\frac{1}{8})),
\end{aligned}
\end{equation}

and $\beta_3(t)$ is the cubic B-spline:

\begin{equation}
\beta_3(t)=\displaystyle \sum_{i=0}^{4} \frac{(-1)^i}{3 !}\left(\begin{array}{c}
4 \\
i
\end{array}\right)\left(t+2-i\right)^3 \cdot \varpi\left(t+2-i\right), t \in R,
\end{equation}
where $\varpi(t)$ is the unit step function

\begin{equation}
\varpi(t)= \begin{cases}0, & t<0, \\ 1, & t \geq 0.\end{cases}
\end{equation}

 $S(t)$ comes out from a linear combination of the normalized and the shifted B-splines of the same order. Consequently, $S(t)$ can inherit nearly all the favourable properties of $\beta_3(t)$, including analyticity, central symmetry, local support, and high-order smoothness. Moreover, 
$S(t)$ can directly interpolate the provided data without the need to solve coefficient equations, a capability that B-spline lacks.

The Fourier transform expressions of $S(t)$:
\begin{equation}
\widehat{S}(\omega)=\left(\frac{451}{3}-\frac{512}{3} \cos \frac{\omega}{16}+\frac{64}{3} \cos \frac{\omega}{8}\right)\left(\frac{\sin \frac{\omega}{2}}{\frac{\omega}{2}}\right)^4.
\end{equation}

\begin{figure}[ht]
        \includegraphics[width=0.98\linewidth]{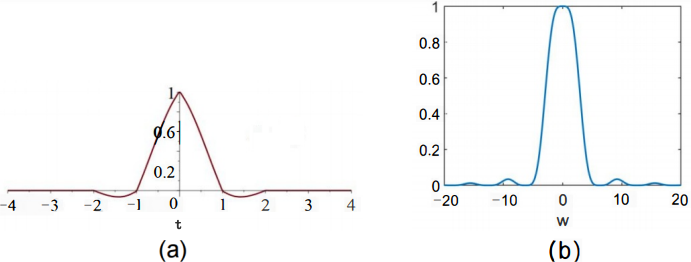}
    \caption{Analysis 
 of the cubic special spline $S(t)$. (a) The graph of the $S(t)$. (b) The graph of the Fourier transform $\widehat{S}(\omega)$. }
    \label{fig:2}
\end{figure}

The spline $S(t)$ and the Fourier transform $\widehat{S}(\omega)$ are separately plotted in Fig.~\ref{fig:2}.

\subsubsection{Constructing Biorthogonal Cubic Special Spline Wavelet (BCSSW)} %MDPI:  %response： %response: I have checked and revised all.
 
\cite{ref-book1} and \cite{ref-journal14} have proved B-spline $\beta_m (t)$ is the scale function of the corresponding multi-resolution analysis. $S(t)$ is formed by the linear combination of B-spline $\beta_3 (t)$ translation and expansion. Therefore, we can naturally deduce the following conclusion.

The subspaces $V_j^3$ are generated by $S(t)$ binary dilation and integer translation, as follows:
\begin{equation}
V_j^3=\overline{\operatorname{span}}\left\{2^{\frac{j}{2}} S\left(2^jt-k\right), k \in Z\right\}, j \in Z,
\label{equ:BGBCSSW1}
\end{equation}
where $\left\{V_j^3\right\}_{j \in \mathbf{Z}}$ forms a general multi-resolution analysis (GMRA) in $L^2(\mathbf{R})$, called spline multi-resolution analysis. $g(t)$ is the corresponding scaling function. According to the theory of wavelet construction, $S(t)$, as a scale function, can construct a new wavelet $\psi(t)$. Let $S^*(t)$ be the dual scaling function of $S(t)$ and  $\psi^*(t)$ be the dual wavelet of $\psi(t)$, then their corresponding low-pass filters are:
\begin{equation}
H(\omega)=\frac{1}{2} \sum_{n=N_1}^{N_2} h_n \mathrm{e}^{-\mathrm{i} n \omega}, \quad H^*(\omega)=\frac{1}{2} \sum_{n=L_1}^{L_2} h_n^* \mathrm{e}^{-\mathrm{i} n \omega}.
\end{equation}

And high-pass filters are:
\begin{equation}
G(\omega)=\frac{1}{2} \sum_{k=1-L_2}^{1-L_1} g_k \mathrm{e}^{-\mathrm{i} k \omega}, \quad G^*(\omega)=\frac{1}{2} \sum_{k=1-N_2}^{1-N_1} g_k^* \mathrm{e}^{-\mathrm{i} k \omega},
\end{equation}
where $N_1, N_2, L_1, L_2$ are all integers, $N_2-N_1+1$ and $L_2-L_1+1$ are the lengths of $H(\omega)$ and $H^*(\omega)$, respectively, and $g_k=(-1)^k h_{1-k}$, $g^*_k=(-1)^k h^*_{1-k}$. All coefficients are real coefficients. 

We also construct a new class of compactly supported wavelets based on CDF. We are aware that wavelets with compact supports exist as long as the two-scale sequence of the related scaling function is finite. In the paper, we set $H(\omega)$ and $H^*(\omega)$ as %please ensure intended meaning is retained
 odd-length and the support set is symmetric at %please ensure intended meaning is retained
 about 0. The vanishing moment order 
of $H(\omega)$ and $H^*(\omega)$ are $N$ and $N^*$, respectively, and they can also have the following representation:
\begin{equation}
H(\omega)=cos(\frac{\omega}{2})^{2N}Q(cos(\omega)),
\label{eq2}
\end{equation}

\begin{equation}
\label{eq3}
H^*(\omega)=cos(\frac{\omega}{2})^{2N^*}Q^*(cos(\omega)),
\end{equation}
where $Q(cos(\omega))$, $Q^*(cos(\omega))$ are the polynomials of $cos(\omega)$.

Let
\begin{equation}
\label{eq7}
P(sin^2(\displaystyle\frac{\omega}{2}))=Q(cos(\omega)) \overline{Q^*(cos(\omega))}, 
\end{equation}
and when $y=sin^2(\displaystyle\frac{\omega}{2})$,
we also have:
\begin{equation}
P(y)= \sum_{n=0}^{L-1}\binom{L-1+n}{n}y^n,
\label{eq4}
\end{equation}
where $L=N+N^*$.

From the time domain expression of the two-scale equation corresponding to $S(t)$, the low-pass filter of $\psi(t)$ in the frequency domain can be obtained as follows:
\begin{equation}
\label{eq5}
H(\omega)=\frac{\widehat{S}(2\omega)}{\widehat{S}(\omega)}
=\frac{451-512 cos\displaystyle \frac{\omega}{8}+64 cos\displaystyle \frac{\omega}{4}}{451-512cos\displaystyle \frac{\omega}{16}+64 cos\displaystyle \frac{\omega}{8}}cos^4(\frac{\omega}{2}).
\end{equation}

From (\ref{eq2}), (\ref{eq7}), and (\ref{eq5}), we can obtain $N=2$, and
\begin{equation}
Q(cos(\omega))=\frac{451-512cos\displaystyle\frac{\omega}{8}+64cos\displaystyle\frac{\omega}{4}}{451-512 cos\displaystyle\frac{\omega}{16}+64cos\displaystyle\frac{\omega}{8}},
\end{equation}
\begin{equation}
Q^*(cos(\omega))=\overline{\frac{P(sin^2(\displaystyle\frac{\omega}{2}))}{Q(cos(\omega))}}.
\label{eq8}    
\end{equation}

When $N=2$, $L$ takes different values, for example, $L=4,5,6,7$ , we can obtain multiple corresponding $N^*$. Bring these values into Equations (\ref{eq4}), (\ref{eq3}), and (\ref{eq8}), by taking in the inverse Fourier transform, we can obtain multiple groups $h_n, h^*_n$ of the corresponding low-pass filter coefficients of the new biorthogonal spline wavelet. Considering the symmetry of the coefficients, we only give $n=0,1,2,3,... $. In practical application, the corresponding odd coefficients can be symmetrically selected for image processing.

We can calculate the corresponding $g_n,g^*_n$, the high-pass filter coefficients of the $\psi(t)$ and $\psi^*(t)$. The filter bank in the frequency domain is $\left\{H(\omega), G(\omega), H^*(\omega), G^*(\omega)\right\}$, the decomposition and reconstruction processes use two different sets of filters, respectively. It was decomposed with $\left\{h^*_n\right\}$ and $\left\{g^*_n\right\}$, the reconstruction uses a different pair of filters $\left\{h_n\right\}$ and $\left\{g_n\right\}$. Because of this, we make $\left\{h^*_n\right\}$ and $\left\{g^*_n\right\}$ wavelet decomposition filters, and $\left\{h_n\right\}$ and $\left\{g_n\right\}$ wavelet synthesis filters.

\subsection{Modulus maxima based on structural uncertainty-aware perception}
Essentially, structural uncertainty perception is a global feature filter for structural uncertainty. Specifically, the algorithm consists of three parts: structural uncertainty feature selection, wavelet-based modulus maxima, and adaptive threshold filter. The pseudo-code for the algorithm is given in Algorithm~\ref{alg:cap}.

\subsubsection{Structural uncertainty-aware feature selector}
The algorithm analyzes the structural statistics of the low-frequency $cA$, including the mean $\mu_{D}$ and standard deviation $\delta_{D}$. The acceptable deviation range is set to 0.05, which corresponds to a permissible detection error of 5\%. Each standard deviation of each high-frequency component modulus $\delta_{D}(h)$, $\delta_{D}(v)$, $\delta_{D}(d)$ is calculated. The low-frequency component distribution $N_a \sim \mathcal{N}(\mu_{D}, \delta_{D})$, is defined as the candidate edge information. The deviation of the standard deviation is then calculated to determine if it satisfies the error range. This is equivalent to converting the detection deviation of the edge to a global threshold setting, which determines the final selected region of the true detection edge.

\begin{algorithm}
\caption{Structural uncertainty-aware feature selection}\label{alg:cap}
\begin{algorithmic}
\REQUIRE Input: $cA$, $cH$, $cV$, $cD$
\ENSURE Output: Edge images $CH'$, $CV'$, $CD'$ of size ($H \times W$)
\STATE $T \gets 0.05$
\STATE $B \gets [7, 7]$
\STATE $S_T \gets B/2$
\FOR{$1 \leq i \leq W-B(1)+1$ and $1 \leq j \leq H-B(2)+1$}
    \STATE Compute $\delta_{D}$, $\delta_{D}(h)$, $\delta_{D}(v)$, $\delta_{D}(d)$, and $\mu_{E}$ of each $B_i$.
    \IF{$N_a \sim \mathcal{N}(\mu_{D}, \delta_{D})$ and $\mu_{D} \in (0, T)$ and $\delta_{D}(*) \in (I, I + T)$ and $\delta_{D} - \delta_{D}(*) < T$}
        \STATE Compute local modulus maxima of each $B_i$. See (\ref{equ13}), (\ref{equ15}), (\ref{equ17}).
    \ELSE
        \STATE $CH'$, $CV'$, $CD'$ = 0
    \ENDIF
\ENDFOR
\end{algorithmic}
\end{algorithm}

\subsubsection{Wavelet-based modulus maxima}
For the wavelet-based modulus maxima algorithm, we first calculate the modal value and gradient direction (angle) of the detected wavelet component region, which can be defined as follows: 

\begin{equation}
C_{\mathrm{x}}=\frac{\partial C_{1}(x, y)}{\partial x}
\end{equation}

\begin{equation}
C_{\mathrm{y}}=\frac{\partial C_{1}(x, y)}{\partial y}
\end{equation}

\noindent where $C_{1}(x,y)$ is the wavelet components of the first level of decomposition. $C_{x}$ and $C_{y}$ are the gradients of the different wavelet components in the horizontal and vertical directions. Then determine the neighborhood coordinates of the pixel based on the angle.
For the low-frequency modulus of the second branch $M_{\mathrm{u}}(x, y)$ can be expressed as follows:

\begin{equation}
\label{equ13}
M_{\mathrm{u}}(x, y)=\sqrt{\left|C_{x}\right|^{2}+\left|C_{y}\right|^{2}}
\end{equation}
where $\lvert C_{x} \rvert$ and $\lvert C_{y} \rvert$ are the modulus components corresponding to the x and y directions. Similarly, the computation of the modal values of the high-frequency components $M_{\mathrm{c}}(x, y)$ is specified as (\ref{equ14}):

\begin{equation}
\label{equ14}
M_{\mathrm{c}}(x, y)=\sqrt{\left|C_{H}\right|^{2}+\left|C_{V}\right|^{2}+\left|C_{D}\right|^{2}}
\end{equation}

where $\lvert C_{H}(x,y)\rvert$, $\lvert C_{V}(x,y)\rvert$, $\lvert C_{D}(x,y)\rvert$ denote the modulus of the horizontal, vertical and diagonal components respectively. The direction of the separate wavelet components $A_{u}$ can be shown in (\ref{equ15}):

\begin{equation}
\label{equ15}
A_{u}=\arctan \left(\frac{C_{y}}{C_{x}}\right)
\end{equation}
\noindent Similarly, the direction of the high-frequency components $A_{s}$ is shown in Equation (6):

\begin{equation}
A_{s}=\arctan \left(\frac{C_{H}}{C_{V}}\right)
\end{equation}

Specifically, we choose $\frac{\pi}{4}$ and $\frac{3\pi}{4}$ to determine whether there is an approximate edge gradient direction and obtain the corresponding neighbor coordinates. Comparing the modulus of the two neighboring points in the gradient direction, a pixel is considered locally maximal if it is the neighborhood maximum in the gradient direction, while the other edge pixels are 0. The modulus maxima $M'(x, y)$ can be expressed as follows:

\begin{equation}
\label{equ17}
M'(x, y) = \begin{cases}
M(x, y), & \begin{aligned}
    \text{if } & M(x, y) > M(n_{x1}, n_{y1}) \text{ and }\\
    & M(x, y) > M(n_{x2}, n_{y2})
\end{aligned} \\
0, & \text{otherwise}
\end{cases}
\end{equation}

\noindent where $M\left(n_{x 2}, n_{y 2}\right)$ denotes the modulus of M(x, y-1), M(x, y+1), and $M\left(n_{x 1}, n_{y 1}\right)$ denotes the modulus of M(x-1, y-1), M(x+1, y+1). This ensures that only local maxima along the gradient direction are retained to effectively sparse the edges, and non-zero values are used as the final matrix of edge coefficients.

\subsubsection{Adaptive threshold filter}
To better adapt to differences in edge strengths across gradient directions, we apply an adaptive threshold filter to the edge strengths using the average of the maximum modal value $D_{\max}$ and the minimum modal value $D_{\min}$ as the threshold value $T$. This produces the final edge image ($M_{f}$), with the threshold value $T$ determined by (\ref{equ18}):

\begin{equation}
\label{equ18}
T=\left(D_{\max }+D_{\min }\right) / 2
\end{equation}

\subsection{Morphology detection based on the multi-structure anti-noise operator}
For the design of multi-structural anti-noise morphological operators, we combine structural elements in three various directions to comprehensively consider texture information in each direction of the detected object. We denote g(x) as the input gray-scale image and the structural elements are denoted as $\lambda(x)$. Assuming that $\beta$ belongs to $\mathcal{R}$, a 3 $\times$ 3 matrix, considering the uniformity of the response intensity in different directions of the edges and the orientation of the wavelet components to efficiently detect the edges in all directions, we design three sets of structural elements with the bases respectively in (\ref{equ19}):

\begin{equation}
\label{equ19}
\begin{aligned}
\lambda_{1}=\mu \begin{bmatrix}0.5 & 1 & 0.5 \\1 & 2 & 1 \\0.5 & 1 & 0.5\end{bmatrix}, \\
\lambda_{2}=\mu \begin{bmatrix}0 & 0.5 & 0 \\0.5 & 0.5 & 0.5 \\0 & 0.5 & 0\end{bmatrix}, \\
\lambda_{3}=\mu \begin{bmatrix}0.5 & 0 & 0.5 \\0 & 0.5 & 0 \\0.5 & 0 & 0.5\end{bmatrix}, \\
\lambda_{h}= \begin{bmatrix}-1 & -1 & -1 \\-1 & 8 & -1 \\-1 & -1 & -1\end{bmatrix}
\end{aligned}
\end{equation}

\noindent where the weight $\mu$ amplifies the intensity to adjust the region brightness. The value of $\mu$ is 2 and \(\lambda_{h}\)  is the Laplacian operator. \par

Dilation increases the size of the foreground by calculating the maximum value within the region of structural elements. Whereas erosion aims to refine the edges of the foreground, morphological dilation, and erosion can be shown as follows:

\begin{equation}
(\mathrm{g} \Theta \lambda)(x)=\min _{y \in \beta}(g(y)-\lambda(y-x)) 
\end{equation}

\begin{equation}
(\mathrm{g} \oplus \lambda)(x)=\max _{y \in \beta}(g(y)+\lambda(x-y)) 
\end{equation}

where (x-y) $\in$ $\mathcal{R}$, $\beta$ is the domain of the structural elements, and $\mathcal{R}$ is the domain of the gray-scale image. \par

Opening operation can preserve the shape structure and size information of the low-frequency components, combined with the structural elements of the design to remove small objects and noise, which is represented as follows:
\begin{equation}
(g \circ \lambda)=(\mathrm{g} \Theta \lambda) \oplus \lambda
\end{equation}

Finally, the multi-structure anti-noise operator is expressed as follows:
\begin{equation}
E_{d}=\left[\left(g \oplus \lambda_{1} \Theta \lambda_{2}\right) \circ \lambda_{3}\right]-\left[\left(g \oplus \lambda_{1} \Theta \lambda_{2}\right) \Theta \lambda_{3}\right]
\end{equation}

\noindent where $E_{d}$ denotes the edge image obtained by the multi-structure anti-noise morphological operator. This operator suppresses edge noise effectively, smooths the boundaries of larger objects, and retains important structural information by utilizing the difference in secondary corrosion.

\subsection{Fusion strategies for morphological refinement and reconstruction}
The morphological refinement and reconstruction fusion enhances the edges, making them easier to distinguish from the background. This process ensures that the final edge map has more continuous and coherent edges, resulting in a more accurate depiction of the area to be detected. The first level of decomposition preserves clear texture in the image, while the second and third branch reconstruction captures the image's structure more accurately.  The first branch fusion is accomplished using the direct weighted fusion method. We utilized the fused edges of the wavelet decomposition of the first branch as a mask to align with the wavelet reconstructed edges, which improved the detection accuracy.
The edge refinement output $F_{d}$ can be expressed as follows:
\begin{equation}
F_{d}=\left(\mathrm{g} \oplus \lambda_{h}\right)-\left(\mathrm{g} \Theta \lambda_{h}\right) 
\end{equation}

\noindent $\mathrm{G}_{i}$ obtained by morphological reconstruction is:
\begin{equation}
\mathrm{G}_{i}=\min \left(E\left(g_{i-1}\right), E_{m}\right) 
\end{equation}

\noindent where $E_{m}$ denotes mask, $E(\cdot)$ indicates erosion, $\mathrm{G}_{i}$ denotes edge after morphological remodeling, and $g_{i-1}$ denotes markers for morphological refinement. \par
Finally, we perform a context-weighted fusion of the mask $E_{m}$ and the reconstructed edge $\mathrm{G}_{i}$ to obtain the final edge detection image $E_{u}$, which can be expressed as follows:

\begin{equation}
E_{\mathrm{u}}=\alpha G_{\mathrm{i}}+(1-\alpha) E_{m}
\end{equation}

\noindent where $\alpha$ is 0.7. Similarly, the mask $E_{m}$ is obtained by weighted fusion of $E_{h}$ and $E_{l}$ via Equation (16) as well.
\begin{figure*}[ht]
\centering
\includegraphics[scale=0.12]{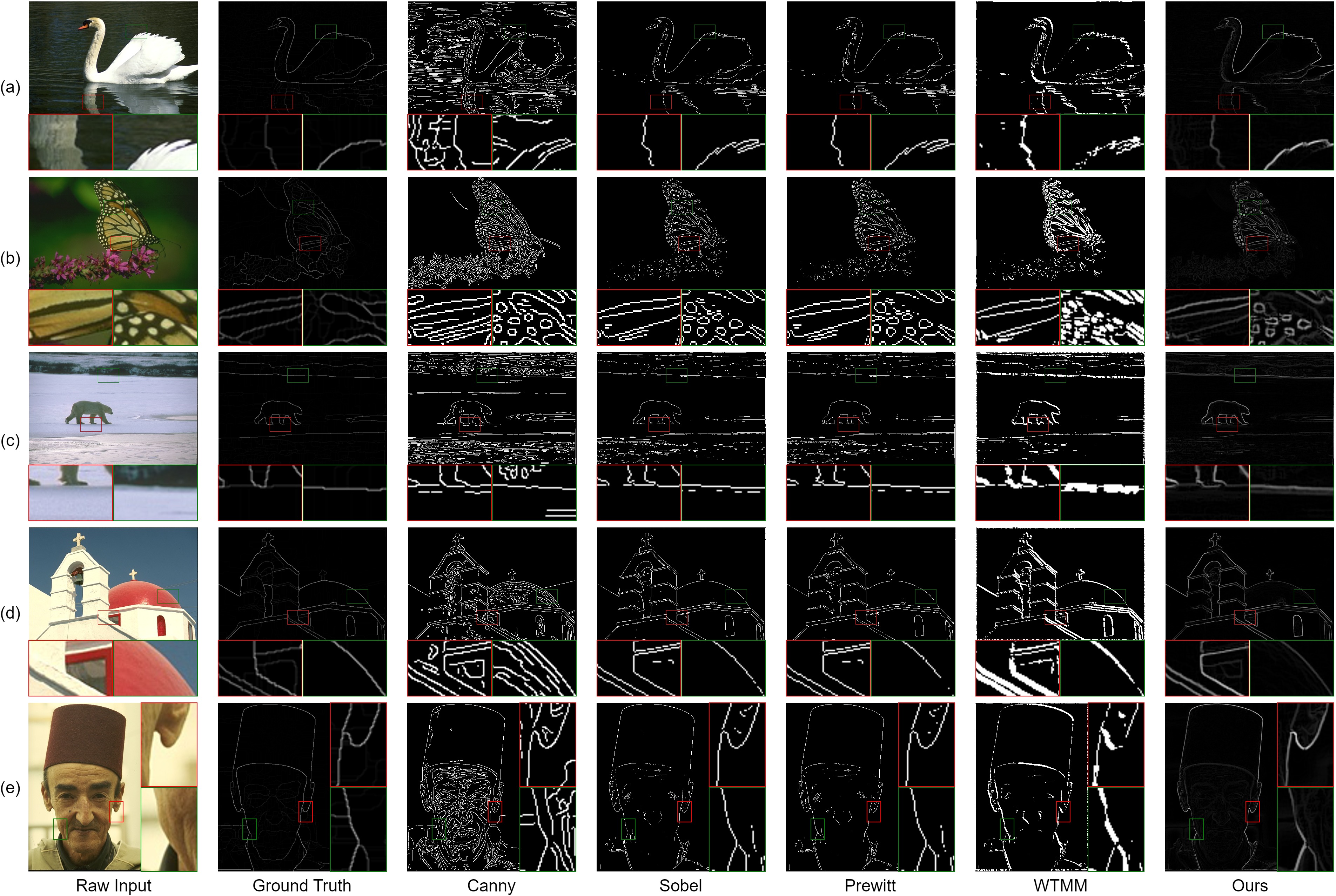}
\caption{Qualitative comparison results on selected samples in BSDS500 dataset with different wavelet detection.}
\label{fig2}
\end{figure*}

\section{Experiments}
In order to demonstrate the effectiveness of the proposed spline wavelet and edge detection algorithm framework, we selected images from BSDS500 and MVTec datasets with labeled datasets for edge detection. We compared and analyzed the results with traditional detection operators such as Canny, Sobel, Prewitt, and the traditional wavelet transform modulus maxima (WTMM) while deploying different wavelets on the proposed algorithm, including Haar, DB2, coif1, rbio3.5, and sym4.
\subsection{Evaluation Metrics}
We consider MSE, PSNR, SSIM, and Entropy as metrics to evaluate the effectiveness of the proposed algorithm in detecting and handling noise with different wavelets. The images are converted to gray-scale before input to evaluate detection effectiveness better.

\subsection{Qualitative evaluation}
For evaluation, we select images from BSDS500 that represent various real-life environments. The results of edge detection are shown in Fig.~\ref{fig2}. Traditional operators tend to detect a significant number of false, noisy, and discontinuous edges, which are easily affected by noise and less robust. On the other hand, the Canny operator faces difficulties in adapting to edge reconstruction and differentiating the main edges. For Sobel, Prewitt, and WTMM operators, Figure (c) clearly shows that there is a problem with complete structural representation, and the results of edge detection do not reflect the reconstruction effect of human visual perception well. The detected edges are unable to distinguish the main objects well, and the edges are rough with imprecise and unnatural edge textures as seen in Figure (a), (b), and (c). Our method is more textured and less sensitive to noise. It presents edge continuity and smoother lines and removes secondary edges from the background. By refining the local edge details, it is evident that the edges reconstructed by the BCSSW are finer and contain more natural details for the main object structures.

To demonstrate the generalization ability of the proposed operator, we also perform tests on the MVTec dataset, which include industrial gray-scale maps and images with varying light intensities. As depicted in Fig.~\ref{fig3}, the conventional operator exhibited high sensitivity to low-light and blurred industrial images, as well as noisy backgrounds. The Canny operator is not effective in detecting partial edges and produces a significant amount of noise. While Sobel, Prewitt, and WTMM reduce some of the extraneous noise, there are still many spurious edges in the interior of the part. The structural information is not accurately detected, and there are inhomogeneous chutes and broken edges. In contrast, the proposed method effectively smooths the background, suppresses extraneous noise, and extracts continuous and smooth edges. \par
\begin{table*}[ht]
    \caption{Evaluation Results on Different Samples with Different Existing Operators in Terms of MSE, PSNR, and Entropy on the BSDS500 dataset. The Best Results Are Shown in Bold. \label{tab1}}
    \begin{center}
    \renewcommand{\arraystretch}{1.2}
    \scalebox{0.92}
    {
        \begin{tabular}{c|ccccccccccc}
\hline
Metrics                   & Image & Canny   & Sobel   & Prewitt & WTMM    & edge\_db2 & edge\_rbio3.5 & edge\_coif1 & edge\_sym4 & edge\_haar      & Ours         \\ \hline
\multirow{5}{*}{MSE}      & (a)   & 0.0965 & 0.0188 & 0.0187 & 0.0389 & 0.0025   & 0.0025       & 0.0022     & 0.0025    & 0.0023         & \textbf{0.0022} \\
                          & (b)   & 0.0703 & 0.0257 & 0.0257 & 0.0578 & 0.0040   & 0.0044       & 0.0040     & 0.0041    & 0.0040         & \textbf{0.0039} \\
                          & (c)   & 0.0914 & 0.0348 & 0.0350 & 0.0566 & 0.0028   & 0.0034       & 0.0026     & 0.0029    & 0.0028         & \textbf{0.0025} \\
                          & (d)   & 0.0567 & 0.0257 & 0.0260 & 0.0741 & 0.0039   & 0.0045       & 0.0040     & 0.0037    & 0.0038         & \textbf{0.0037} \\
                          & (e)   & 0.0694 & 0.0167 & 0.0167 & 0.0413 & 0.0034   & 0.0036       & 0.0034     & 0.0034    & 0.0035         & \textbf{0.0033} \\ \hline
\multirow{5}{*}{PSNR(dB)} & (a)   & 10.1543 & 17.2579 & 17.2739 & 14.1013 & 26.0812   & 25.9929       & 26.5446     & 26.0820    & 26.4043         & \textbf{26.6327} \\
                          & (b)   & 11.5335 & 15.9090 & 15.9060 & 12.3837 & 23.9802   & 23.5745       & 23.9923     & 23.8794    & 23.9714         & \textbf{24.1469} \\
                          & (c)   & 10.3909 & 14.5838 & 14.5648 & 12.4760 & 25.5107   & 24.6628       & 25.8221     & 25.3203    & 25.6113         & \textbf{26.0865} \\
                          & (d)   & 12.4641 & 15.9051 & 15.8556 & 11.3009 & 24.0882   & 23.4775       & 23.9980     & 24.3070    & 24.1931         & \textbf{24.3799} \\
                          & (e)   & 11.5847 & 17.7053 & 17.7836 & 13.8404 & 24.7299   & 24.4486       & 24.6468     & 24.6891    & 24.5442         & \textbf{24.8339} \\ \hline
\multirow{5}{*}{Entropy}  & (a)   & 0.4778  & 0.1541  & 0.1539  & 0.2630  & 0.9342    & 0.9148        & 0.9405      & 0.9389     & 0.9308          & \textbf{0.9460}  \\
                          & (b)   & 0.3909  & 0.1811  & 0.1814  & 0.3362  & 0.9653    & 0.9603        & 0.9786      & 0.9657     & \textbf{0.9975} & 0.9784           \\
                          & (c)   & 0.4552  & 0.2280  & 0.2288  & 0.3377  & 0.9028    & 0.8814        & 0.9092      & 0.9112     & 0.8962          & \textbf{0.9154}  \\
                          & (d)   & 0.3464  & 0.2022  & 0.2035  & 0.4240  & 0.9736    & 0.9872        & 0.9679      & 0.9610     & 0.9336          & \textbf{0.9670}  \\
                          & (e)   & 0.3868  & 0.1409  & 0.1391  & 0.2810  & 0.8895    & 0.8678        & 0.9030      & 0.8954     & \textbf{0.9253} & 0.9072           \\ \hline
\end{tabular}
    }
    \end{center}
\end{table*}

\begin{figure*}[ht]
\centering
\includegraphics[scale=0.16]{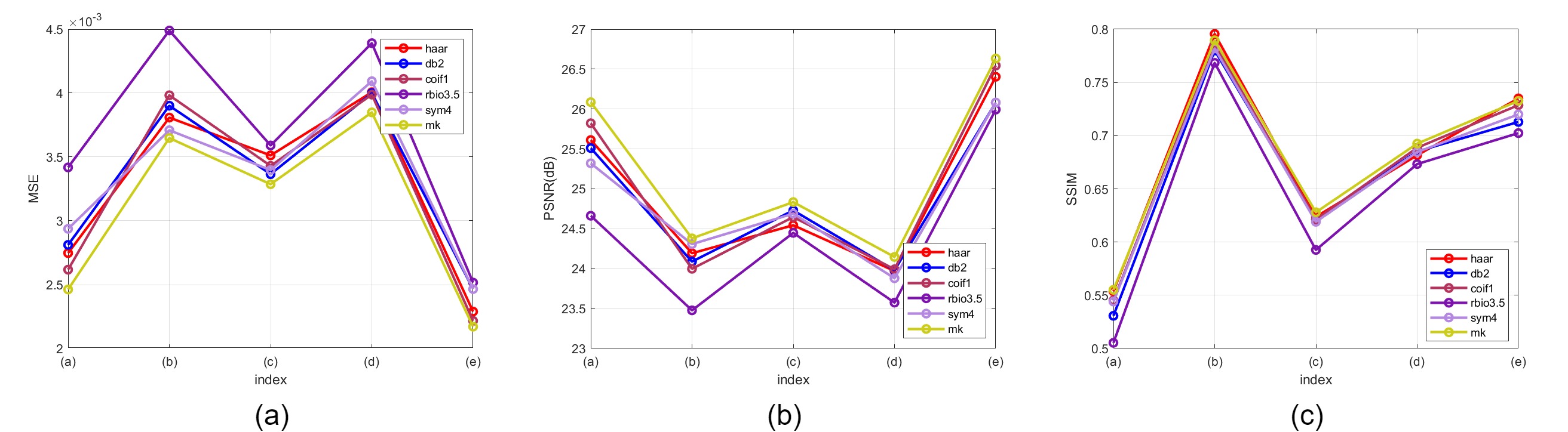}
\caption{Fold analysis results on selected samples in BSDS500 dataset with different wavelet detection.}
\label{fig5}
\end{figure*}

\begin{table}[ht]
    \caption{Evaluation Results on Detected Samples Using Different Wavelets in Terms of SSIM on BSDS500. The Best Results Are Shown in Bold. \label{tab2}}
    \begin{center}
        \scalebox{0.8}{
        \begin{tabular}{c|ccccccc}
\hline
Metric                & Image & DB2    & rbio3.5 & coif1  & sym4   & Haar   & Ours   \\ \hline
\multirow{5}{*}{SSIM} & (a)   & 0.7129 & 0.7024  & 0.7287 & 0.7200 & \textbf{0.7350} & 0.7327 \\
                      & (b)   & 0.6852 & 0.6733  & 0.6886 & 0.6846 & 0.6816 & \textbf{0.6925} \\
                      & (c)   & 0.5307 & 0.5053  & 0.5452 & 0.5441 & 0.5537 & \textbf{0.5551} \\
                      & (d)   & 0.7800 & 0.7684  & 0.7855 & 0.7820 & \textbf{0.7956} & 0.7896 \\
                      & (e)   & 0.6209 & 0.5926  & 0.6212 & 0.6190 & 0.6237 & \textbf{0.6280} \\ \hline
\end{tabular}
        }
    \end{center}
\end{table}

\begin{table}[ht]
    \caption{Evaluation Results on Detected Samples Using Different Wavelets in Terms of SSIM on BSDS500. The Best Results Are Shown in Bold. \label{tab3}}
    \begin{center}
        \scalebox{0.8}{
        \begin{tabular}{c|cccccc}
\hline
Metric                   & Image     & canny  & sobel  & prewitt & WTMM   & Ours              \\ \hline
\multirow{5}{*}{Entropy} & (f)       & 0.7211 & 0.1580 & 0.1560  & 0.2047 & \textbf{0.9817} \\
                         & (g)       & 0.6757 & 0.1353 & 0.1327  & 0.2616 & \textbf{0.9761} \\
                         & normal(h) & 0.1450 & 0.0583 & 0.0584  & 0.0979 & \textbf{0.9975} \\
                         & bright(i) & 0.4382 & 0.0567 & 0.0571  & 0.0878 & \textbf{0.9766} \\
                         & dark(j)   & 0.2500 & 0.0649 & 0.0644  & 0.0726 & \textbf{0.9185} \\ \hline
\end{tabular}
        }
    \end{center}
\end{table}

To further demonstrate the robustness of the detection operators to the light-generated noise, Fig.~\ref{fig4} shows the results for industrial images under different lighting scenarios, where the Canny operator also exhibits light-sensitive properties, and the other detection operators embody the robustness to background noise to a certain extent, the edges of the detected object's structure are poorly preserved, and there are discontinuities and varying degrees of background noise texture. The proposed algorithm, on the other hand, can maximize the suppression of the generation of background noise due to light intensity variations, while extracting smooth and meaningful edge information. This further illustrates the generalization and robustness of the proposed algorithm in various detection scenarios.

\subsection{Quantitatively Evaluation}
Due to its subjective structural description caused by human labeling of edge differences, we choose MSE, PSNR, and Entropy to quantitatively analyze the effect of edge detection to assess the real edges and perceived structural representations. A smaller MSE (or larger PSNR) indicates that the edge image contains less noise and the detection result retains more effective edge content. We aim for a reasonable entropy to reflect the intensity distribution and detailed representation of edges. Fig.~\ref{fig5} shows the fold analysis of different wavelet edge detection results. Table.~\ref{tab1} shows that BCSSW outperforms other existing wavelets in MSE and PSNR values and is significantly more robust to noise than the traditional detection operator. The entropy is highly competitive and the plots in (a), (b), and (e) indicate that the extracted edge strength is smoother and more effective in edge information detection compared to traditional operators. Table.~\ref{tab2} shows that the SSIM values of the wavelet detected images are higher than those of other existing wavelets, indicating that BCSSW effectively preserves the true edge structure and can smooth the image while maintaining noise robustness. In addition, Table.~\ref{tab3} presents a comparison of entropy results for different detection operators on the MVTec dataset. The entropy of Sobel, Prewitt, and WTMM for the image (h) (i) (j) are close to 0, indicating an unbalanced distribution of edge intensities, which is insufficient to represent the complex structural edges, and the proposed algorithm obtains the highest values for detection, which further illustrates the generalization ability of this algorithm in different light intensity and noise environments. The proposed algorithm achieves the highest detection values, demonstrating its ability to generalize across different light intensities and noise environments.

\begin{figure*}[ht]
\centering
\includegraphics[scale=0.06]{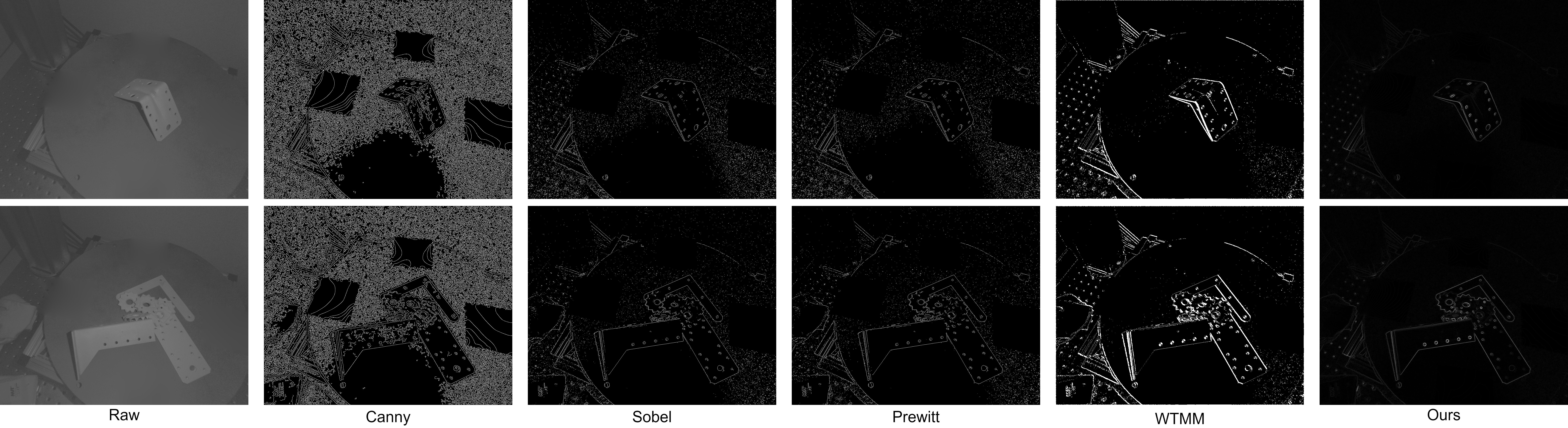}
\caption{Qualitative comparison results on industrial samples in MVTec ITODD dataset with different operator detection.}
\label{fig3}
\end{figure*}

\begin{figure*}[ht]
\centering
\includegraphics[scale=0.04]{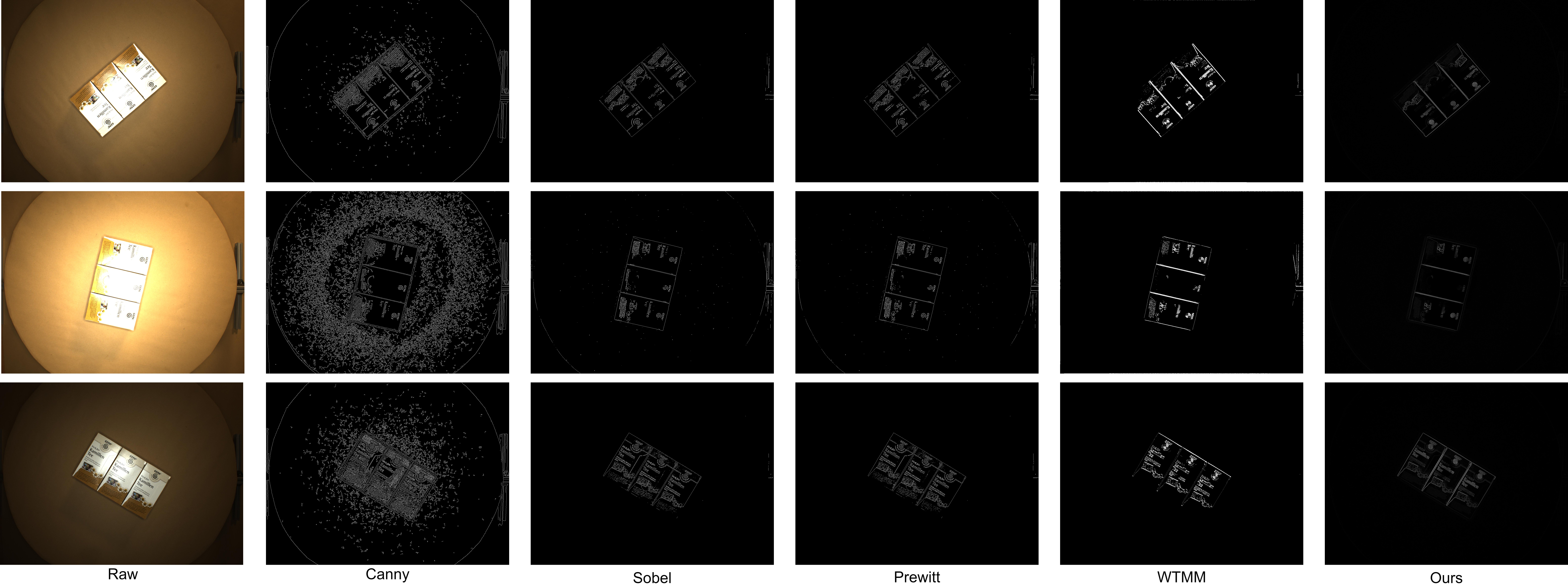}
\caption{Qualitative comparison results on industrial samples in MVTec D2S dataset with different operator detection.}
\label{fig4}
\end{figure*}

\begin{table}[ht]
    \caption{Ablation Results for the Proposed Algorithm. \label{tab4}}
    \begin{center}
        \scalebox{0.8}{
        \begin{tabular}{c|ccccc}
\hline
Metrics                   & Image & w/o I           & w/o III & w/o I,II        & Ours\_         \\ \hline
\multirow{5}{*}{MSE}      & (a)   & 0.00309         & 0.00269 & 0.00297         & \textbf{0.00217} \\
                          & (b)   & 0.00452         & 0.00470 & 0.00426         & \textbf{0.00385} \\
                          & (c)   & 0.00294         & 0.00372 & 0.00269         & \textbf{0.00246} \\
                          & (d)   & 0.00435         & 0.00452 & 0.00414         & \textbf{0.00365} \\
                          & (e)   & 0.00357         & 0.00399 & 0.00348         & \textbf{0.00329} \\ \hline
\multirow{5}{*}{PSNR(dB)} & (a)   & 25.0963         & 25.6984 & 25.2739         & \textbf{26.6327} \\
                          & (b)   & 23.4523         & 23.2782 & 23.7045         & \textbf{24.1469} \\
                          & (c)   & 25.3116         & 24.2950 & 25.7072         & \textbf{26.0865} \\
                          & (d)   & 23.6162         & 23.4536 & 23.8266         & \textbf{24.3799} \\
                          & (e)   & 24.4777         & 23.9892 & 24.5893         & \textbf{24.8339} \\ \hline
\multirow{5}{*}{SSIM}     & (a)   & 0.7526          & 0.6239  & \textbf{0.7529} & 0.7327           \\
                          & (b)   & 0.6927          & 0.6645  & \textbf{0.6934} & 0.6925           \\
                          & (c)   & 0.7523          & 0.4459  & \textbf{0.7532} & 0.5551           \\
                          & (d)   & 0.8052          & 0.7595  & \textbf{0.8058} & 0.7896           \\
                          & (e)   & \textbf{0.7653} & 0.5609  & 0.7643          & 0.6280           \\ \hline
\end{tabular}
        }
    \end{center}
\end{table}

\subsection{Ablation Study}
In this section, we conduct an ablation study of the structural uncertainty-aware feature selector and the fusion of multi-structural operators to demonstrate the effectiveness of the proposed algorithm. The details are given below:
\begin{itemize}
    \item[-]
    Effectiveness of the structural uncertainty-aware feature selector (w/o III)
    \item[-] 
    Effectiveness of multi-structure anti-noise operator design (w/o I)
    \item[-]
    Effectiveness of operator fusion (w/o I, II)
\end{itemize}

Table.~\ref{tab4} shows that w/o I, II has the least decrease in metrics, while the SSIM values are the highest. In contrast, w/o III has significantly lower SSIM values, reflecting the ability of the structural uncertainty feature selection to better filter out meaningful structural edge representations. The absence of separate multi-structural anti-noise operators results in the lowest MSE and PSNR rankings, demonstrating the contribution of multi-structural operator fusion to improving anti-noise capability and suppressing edge structure to a certain degree. \par

\begin{figure*}[ht]
\centering
\includegraphics[scale=0.18]{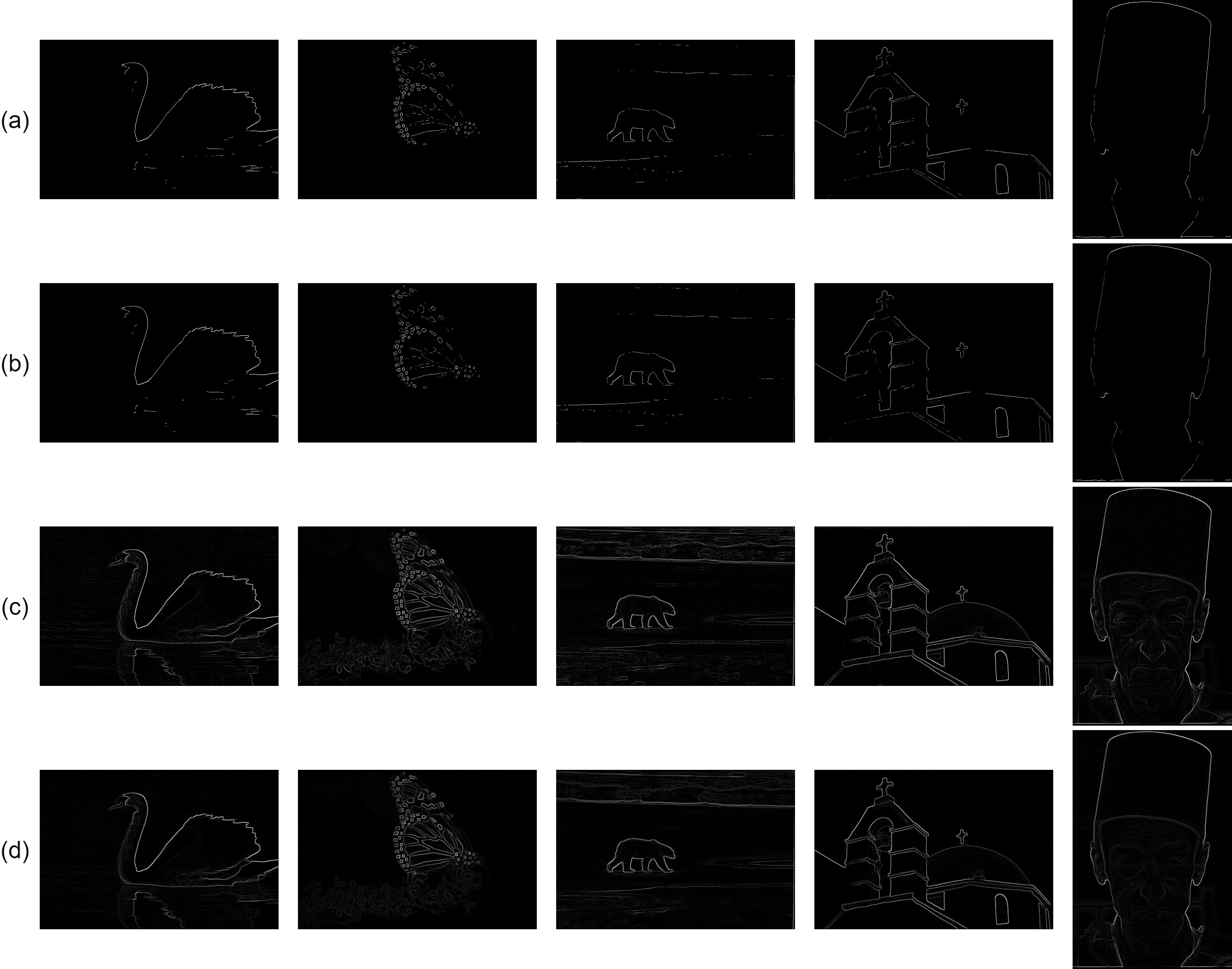}
\caption{Qualitative comparison of the ablation study. (a) is w/o I. (b) is w/o I, II. (c) is w/o III. (d) is Ours.}
\label{fig6}
\end{figure*}

Fig.~\ref{fig6} shows that the detected edges experience edge detail loss and breakage when the multi-structural anti-noise operator and fusion are missing, while in the lack of structural uncertainty-aware feature selection, the edge results present too much detailed texture and even irrelevant structural features. Figure (d) reasonably maintains both the main structural edge information as well as complete and continuous edges. This well illustrates the importance of resultant uncertainty feature selection and multi-structure operator fusion for edge extraction, further confirming the conclusion of the quantitative analysis.

\section{Conclusion}
In this study, we propose a new BCSSW edge detector (EDBSW) based on structural uncertainty perception and anti-noise operator fusion. We leverage the BCSSW to enhance edge detection by perceiving structural uncertainty. This approach guides the detection of modulus maxima, allowing for the extraction of edges with consistent information. By fusing anti-noise morphology operators, we effectively suppress noisy boundaries. The strategy is further refined through a reconstruction process, which enables the extraction of edges that are both resistant to noise and rich in detail. The experimental results further demonstrate the superiority of the proposed detection algorithm in noise robustness compared to other operators and different wavelets and verify the effectiveness in retaining the structural information of edges. Also, the proposed detection algorithm passes the validation of the ablation part. However, our detected edges still have low efficiency and limited generalization. Our future work will focus on constructing a generalized and concise spline wavelet and neural detection operators.

\end{document}